# Data Quality Challenges in Retrieval-Augmented Generation

*Completed Research Paper*

**Leopold Müller**[*]
University of Bayreuth
Bayreuth, Germany
leopold.mueller@uni-bayreuth.de

**Joshua Holstein**[*]
Karlsruhe Institute of Technology
Karlsruhe, Germany
joshua.holstein@kit.edu

**Sarah Bause**[*]
Karlsruhe Institute of Technology
Karlsruhe, Germany
sarah.bause@student.kit.edu

**Gerhard Satzger**
Karlsruhe Institute of Technology
Karlsruhe, Germany
gerhard.satzger@kit.edu

**Niklas Kühl**
University of Bayreuth
Bayreuth, Germany
kuehl@uni-bayreuth.de

## Abstract

*Organizations increasingly adopt Retrieval-Augmented Generation (RAG) to enhance Large Language Models with enterprise-specific knowledge. However, current data quality (DQ) frameworks have been primarily developed for static datasets, and only inadequately address the dynamic, multi-stage nature of RAG systems. This study aims to develop DQ dimensions for this new type of AI-based systems. We conduct 16 semi-structured interviews with practitioners of leading IT service companies. Through a qualitative content analysis, we inductively derive 15 distinct DQ dimensions across the four processing stages of RAG systems: data extraction, data transformation, prompt & search, and generation. Our findings reveal that (1) new dimensions have to be added to traditional DQ frameworks to also cover RAG contexts; (2) these new dimensions are concentrated in early RAG steps, suggesting the need for front-loaded quality management strategies, and (3) DQ issues transform and propagate through the RAG pipeline, necessitating a dynamic, step-aware approach to quality management.*

**Keywords:** Retrieval-Augmented Generation, Data Quality, Large Language Models

## Introduction

The adoption of large language models (LLMs) is transforming enterprise operations by supporting employees in automating knowledge-intensive tasks, improving information retrieval, and ultimately enhancing decision-making (Chang et al., 2025). While LLMs demonstrate superior performance across various tasks (Singhal, 2023), their lack of enterprise-specific data often limits their usefulness in organizational contexts, leading to generic or inaccurate outputs (Parthasarathy et al., 2024; Zhao et al., 2024). Therefore, organizations increasingly leverage Retrieval-Augmented Generation (RAG), an approach that complements LLMs with a knowledge base (Lewis et al., 2020). This integration enables

---

[*]Authors contributed equally to this research.

LLMs to access supplementary information leading to more context-aware and ultimately also to more accurate responses (Shone, 2025). As with any application, the effectiveness of RAG systems fundamentally depends on the quality of the data used. However, maintaining that level of quality is challenged by the dynamic nature of corporate information, which is typically continuously revised and expanded (Kockum & Dacre, 2021). As the companies' proprietary knowledge evolves, for example through regular updates of documents or content revisions, curating high-quality datasets for the knowledge base represents a major challenge for the successful adoption of RAG systems. At the same time, these RAG systems typically involve multiple sequential processing steps to transform raw data into accessible pieces of information that can be utilized by the RAG system (Jeong, 2023). This sequential process introduces unique challenges that go beyond established AI systems: As one processing step builds the foundation for the following ones, early errors can propagate through the system, potentially resulting in unreliable or outdated outputs.

Over recent decades, information systems (IS) research has extensively investigated data quality (DQ) across diverse contexts from data-driven decision making (Khong et al., 2023; Strong et al., 1997) to big data (Cai & Zhu, 2015; Janssen et al., 2017) and, more recently, also for AI applications (Edwards, 2024; Zha et al., 2023). In this context, the rise of data-centric AI (Jakubik et al., 2024), leads to an increasing focus on DQ to improve AI systems' performance. This reinforces the relevance of established DQ frameworks with dimensions such as accuracy, timeliness, and consistency (Pipino et al., 2002; Sidi et al., 2012; Wang et al., 2024; Wang & Strong, 1996), for developing strategies to improve the data's quality , and ultimately, also the overall decision quality (Jakubik et al., 2024). However, these frameworks are mostly developed in the context of static datasets and structured databases (Arolfo & Vaisman, 2018; Cai & Zhu, 2015; Saha & Srivastava, 2014; Zhou, Tu, et al., 2024), making them insufficient for addressing the dynamic and multi-step nature of systems like RAG. Therefore, current research still lacks a systematic understanding of the unique DQ challenges introduced by LLMs, and RAG systems in particular. Bridging this gap, research starts investigating DQ in LLM by examining practitioners' views, focusing primarily on what constitutes high-quality training data (Yu et al., 2024). Yet, these efforts largely center on static, pre-training datasets and overlook the diversity and dynamics of enterprise-specific information that is being used for RAG systems. To address this gap, we pose the following research question:

***RQ:*** *What data quality challenges emerge at the distinct steps of Retrieval-Augmented Generation?*

To address this question, we conduct semi-structured interviews with 16 practitioners from two of the world's largest companies in the IT service industry. Following the established methodology of Gioia et al. (2013), we qualitatively analyze these interviews through inductive coding and systematically derive 26 DQ challenges that we map across the sequential processing steps of RAG systems. By doing so, we make a three-fold contribution to IS research and practice: First, we empirically uncover how DQ challenges manifest in real-world RAG systems. Second, we introduce a step-aware perspective on DQ, demonstrating how specific dimensions emerge and propagate across the sequential steps of RAG. Third, we complement DQ theory (Pipino et al., 2002; Wang et al., 2024; Wang & Strong, 1996) in the context of LLMs by revealing that while established DQ dimensions are present in RAG's initial steps, unique challenges emerge dynamically during downstream processing.

# Foundation and Related Work

Several theoretical foundations are relevant to our work. We begin by outlining core concepts and mechanisms of RAG. This is followed by a discussion of DQ dimensions and a review of related approaches that address DQ within RAG.

### *Retrieval-Augmented Generation*

While businesses are adopting LLMs for tasks like content generation and customer service (Hadi et al., 2023), off-the-shelf models often lack the domain- and enterprise-specific knowledge crucial in many contexts (Ling et al., 2024) . Their limited context windows, i.e., the amount of information they can process at once, further restrict the integration of comprehensive proprietary information (Chen et al., 2023). When confronted with unfamiliar queries, LLMs can produce confident but incorrect answers so-called "hallucinations" which pose risks in accuracy-critical environments (Huang et al., 2025). To overcome these limitations, RAG enhances LLMs by enabling real-time retrieval of relevant external knowledge (Lewis et al., 2020). Based on Jeong's (2023) framework, we outline the RAG process in four steps*: Data Extraction*, *Data Transformation*, *Prompt & Search*, and *Generation* (see Figure 1).

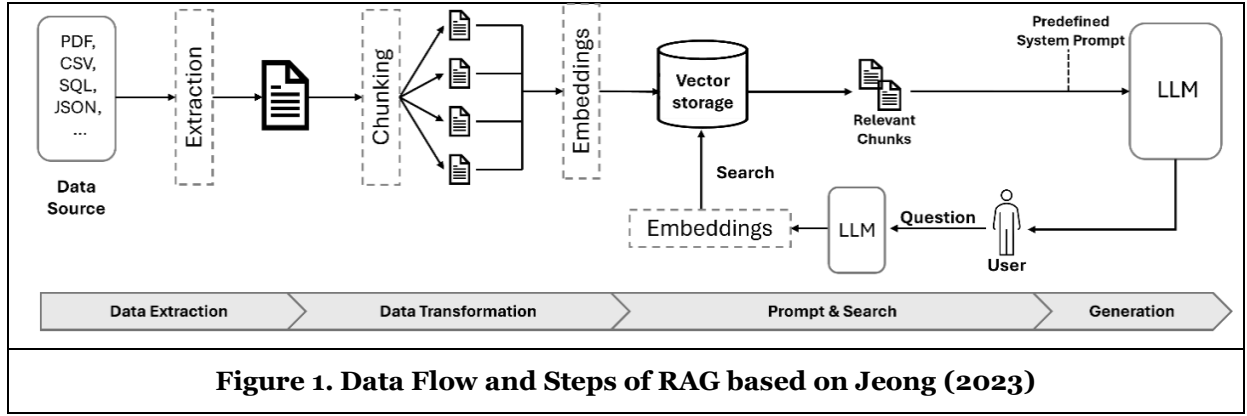

**Figure 1. Data Flow and Steps of RAG based on Jeong (2023)**

*Data Extraction* involves collecting relevant information from various structured and unstructured enterprise data sources (Jeong, 2023). *Data Transformation* then extracts and processes the collected data to make it suitable for LLM processing. As LLMs can only process certain amounts of information at the same time, this transformation step involves segmenting the extracted data into smaller chunks using a chunking strategy. These chunks are then converted into embeddings, indexed, and stored in a vector database for efficient retrieval. During *Prompt & Search*, the user submits a query to the LLM. This query is transformed into an embedding, which is then compared against the embeddings stored in the vector database. Based on similarity matching algorithms, the top N most relevant chunks are retrieved and passed to the generator, ensuring that only the most pertinent information is considered for response generation. In the final *Generation* step, the retrieved data chunks, along with a predefined system prompt that guides the model's behavior and response style, are combined to form the comprehensive prompt. This carefully constructed prompt is provided to the language model, which then synthesizes the information to generate coherent, contextually relevant, and accurate responses that address the user's original query while maintaining alignment with organizational knowledge and constraints.

## *Data Quality*

Enterprise decisions are only as good as the data they rely on—yet in practice, decisions are often made based on incomplete, inconsistent, or otherwise imperfect subsets of available data (Cichy & Rass, 2019). This makes DQ a critical factor in ensuring reliable, effective, and goal-aligned decision-making across organizations. As a result, DQ has become a central topic in information systems research, with numerous frameworks proposed to define, assess, and manage it across contexts (Sidi et al., 2012). While perspectives vary, a widely accepted foundation is the multi-dimensional model introduced by Wang and Strong (1996), which defines DQ in terms of its fitness for use. They group 15 key attributes into four core dimensions: intrinsic DQ (e.g., accuracy, objectivity), contextual DQ (e.g., relevance, timeliness), representational DQ (e.g., interpretability, consistency), and accessibility DQ (e.g., ease of access, security). These dimensions remain highly influential, forming the basis for DQ assessment practices and informing how organizations align data assets with decision-making needs. In practice, DQ assessment typically blends objective measures (such as completeness or consistency) with subjective evaluations (such as understandability or relevance), often guided by frameworks like Total Data Quality Management (Wang, 1998) or AIM Quality (Lee et al., 2002). As organizations increasingly adopt AI-enabled systems, especially those reliant on large-scale, unstructured, or dynamically sourced data, DQ is no longer a static concept evaluated post hoc. It is a continuously evolving property that directly shapes model performance. The current shift toward data-centric AI emphasizes that high-performing models require not just good architecture, but high-quality, context-appropriate data. This perspective makes DQ governance an ongoing operational concern, where data must be monitored, refined, and aligned with changing goals and contexts (Jakubik et al., 2024). These challenges are especially pronounced in multi-step systems like RAG, where DQ must be maintained across distinct but interdependent steps, from retrieving documents to generating outputs. In such systems, established DQ dimensions still provide a valuable foundation, but their operationalization must account for the complexity and fluidity of data use across different steps, prompting a need for more adaptive and context-aware quality management approaches.

### *Data Quality for Retrieval-Augmented Generation*

Substantial research on DQ in the field of IS demonstrates that incorrect or incomplete data can degrade system performance, e.g., Baier et al. (2019) identify various DQ related challenges for implementation of ML pipelines in enterprises, and Sambasivan et al. (2021) identify cascading DQ challenges in ML triggered by factors such as real-world brittleness, knowledge gaps, and metadata issues. Zhou et al. (2024) further identify that opensource DQ tools often only address a very narrow set of DQ metrics, primarily covering the intrinsic, contextual, representational, and accessibility dimensions defined of Wang and Strong (1996). To close this gap, Zhou et al. (2024) propose a more comprehensive metric set and highlight how generative AI augmentation can raise DQ across iterative ML workflows. While these contributions advance DQ practices for ML pipelines, they do not account for the complexity of modern AI systems such as RAG which operate with more dynamic and layered data flows. In the context of RAG, Holstein et al. (2025) demonstrate the importance of metadata for identifying the relevant information. Zhou et al. (2024) shift attention for the evaluation of LLMs by proposing a shift from model accuracy to evaluating its trustworthiness, proposing a benchmark that evaluates RAG systems across six risk dimensions. Jin et al. (2024) introduce a filtering system that uses linguistic heuristics and domain-specific rules to eliminate low-value data while preserving diversity in large-scale datasets. Their findings show that filtered datasets lead to superior model performance, underscoring the value of data curation. Adopting a broader view on DQ, Yu et al. (2024) conduct a mixed methods study to identify 13 dimensions of high-quality data for the pretraining of LLMs. While their findings provide valuable insights into the DQ of static datasets for pretraining, they do not address the dynamic, real-time demands of RAG. Despite growing interest in DQ and RAG, their intersection remains underexplored. Existing research addresses either DQ principles in IS and ML settings, or static pretraining data. What remains missing is a systematic investigation of how DQ challenges apply to RAG. This gap is critical, as RAG introduces new forms of real-time data dependency, layered processing, and error propagation (Barnett et al., 2024; Min et al., 2025). Our research addresses this need by offering a step-aware framework that connects existing DQ theory with the sequential, real-time nature of RAG.

## Methodology

To identify DQ challenges in RAG systems, we conduct semi-structured interviews (Lewis-Beck et al., 2004) with an average duration of 45 minutes. Our interview design draws on established DQ dimensions and RAG process steps (Figure 1) as starting points for questioning, while maintaining openness to novel, step-specific challenges that emerge from participants' experiences. Using purposeful sampling (Patton, 2002), we recruited participants with experience in LLMs, RAG systems, and data management from two of the world's largest IT service companies. Their professional role has allowed the participants to work with various client companies, providing them with a broader understanding of the common DQ challenges emerging in RAG (see Table 1) when developing such systems. All participants were technical practitioners involved in RAG system development, deployment, and implementation. They work on building RAG systems, enabling customer implementations, and consulting on major RAG projects. The experts have worked with RAG since its initial release in the international company's portfolio around 2.5 years ago. This ensures insights reflect practitioners directly engaged in RAG system creation and deployment rather than end-user perspectives. The interviews are recorded and automatically transcribed with Microsoft Copilot followed by manual revision. After 16 interviews, a saturation effect is observed (Patton, 2002), indicating that no additional insights are emerging and the data collection consequently concludes.

In the interviews, participants are first asked whether the high-level RAG model as presented in Figure 1 accurately reflects their perception of RAG systems. All confirm its validity, with minor adjustments suggested for specific processes such as rewriting or reranking, which depend on use case specifics and complexity. In the subsequent analysis of the interviews, we implement an iterative, inductive coding procedure following Gioia et al. (2013) using the terms "concept", "theme", and "dimension" to rank hierarchical categories from specific to more general (Patton, 2002).

| Expert | Role | Industry of customers | # of years in company | # of years in IT industry |
|---|---|---|---|---|
| α | Data Scientist | FSI | > 15 | > 15 |
| β | Data Scientist | AM, EN, HC | 5 – 15 | 5 – 15 |
| γ | Data Scientist | CH, EN, OG | < 5 | 5 – 15 |
| δ | Data Scientist | VA | < 5 | < 5 |
| ε | Cloud Solution Architect | VA | 5 - 15 | > 15 |
| ζ | Cloud Solution Architect | FSI, RT | 5 - 15 | 5 - 15 |
| η | Cloud Solution Architect | AM | 5 - 15 | 5 - 15 |
| θ | Cloud Solution Architect | AM, TT | 5 – 15 | 5 – 15 |
| ι | Cloud Solution Architect | VA | < 5 | 5 – 15 |
| κ | Cloud Solution Architect | CH, LS | < 5 | 5 - 15 |
| λ | Cloud Solution Architect | VA | < 5 | < 5 |
| μ | Consultant | FSI, AM | 5 - 15 | 5 - 15 |
| ν | Consultant | VA | < 5 | > 15 |
| ξ | Consultant | AM | < 5 | < 5 |
| ο | Consultant | CH, HC, LS | < 5 | < 5 |
| π | Cloud Product Manager | AM, FSI, MF | < 5 | 5 - 15 |

*Legend:* AM = automotive; CH = chemicals; EN = energy; FSI = financial service industry; HC = healthcare; LS = life science; MF = manufacturing; OG = Oil & Gas; RT = retail; TT = Transport & Travel; VA = various

**Table 1. List of Interview Experts**

For the analysis, one author performs an open coding of the interview transcripts. Based on the open coding results, we conduct two 3-hour workshops among the authors to create a shared understanding of the codes. In doing so, we develop consensus regarding the concepts identified from the expert interviews and revise our open first-level codes. The author who initially performed the open coding then revised the codes based on insights generated from the first workshop. The revised codes were subsequently discussed in the second workshop, where consensus was reached and no further revisions were deemed necessary. Thereafter, we apply inductive coding as described in Gioia et al. (2013) to connect the first-order concepts and group them into second-order themes. Then, we test these relationships against the data, ultimately improving our understanding of the main second-order themes. In the final step, we further distill the emergent themes into aggregated dimensions as principal categories (Gioia et al., 2013). During the inductive coding procedure of the first step of RAG systems, i.e., the *Data Extraction*, we observe strong content-related similarities between our aggregate dimensions and the dimensions identified by Wang and Strong (1996). This leads us to adopt their established terminology for the challenges' names for consistency and clarity reasons.

## Results

In the following, we present the results of our empirical investigation. Specifically, the DQ challenges identified through the expert interviews are outlined as guided by the applied methodological framework and systematically mapped to the four steps of the RAG.

### *Data Extraction*

In the *Data Extraction* step of the RAG pipeline, relevant information is collected from various enterprise data sources. We identify five DQ challenges: intrinsic, contextual, representational, accessibility, and accountability.

## Intrinsic Data Quality

Intrinsic DQ refers to the inherent characteristics of the data that ensure its accuracy and objectivity (see Figure 2), independent of the context in which it is used. It encompasses the themes Objectivity and Accuracy that are both widely recognized in the field to IS and provide the foundation for evaluating whether data is suitable for further processing and decision-making.

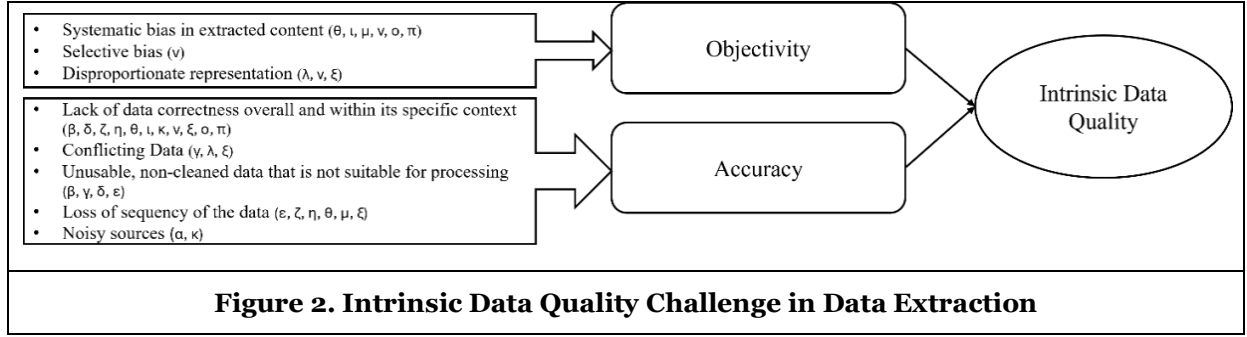

**Figure 2. Intrinsic Data Quality Challenge in Data Extraction**

**Objectivity** captures the extent to which the extracted data is free from bias. Systematic bias refers to recurring patterns of distortion, leading to misinterpretations and incorrect conclusions. In contrast, selective bias occurs when the selection of data sources is non-comprehensive, favoring certain types of documents or perspectives while neglecting others, ultimately, compromising the neutrality of the dataset, as μ states, "disproportionate representation can be problematic, particularly when documents from various business units are collected and merged."

**Accuracy** concerns the correctness and coherence of data in both general and specific contexts. Data may lack correctness either in general or within the specific context in which it is intended to be used. Conflicting data refers to inconsistencies within or across datasets, while noisy data includes irrelevant or low-quality information that obscure meaningful patterns. Finally, when the chronological order of data points is disrupted, interpretation becomes increasingly difficult.

## Contextual Data Quality

Contextual DQ is defined as the extent to which data is relevant, timely, complete, and appropriate for its specific use case. In contrast to intrinsic DQ, which is independent of context, this challenge evaluates how well the data fits the context to which it is applied. As shown in Figure 3, the analysis reveals four key themes: Uniqueness, Coverage, Timeliness and Versioning, and Relevance.

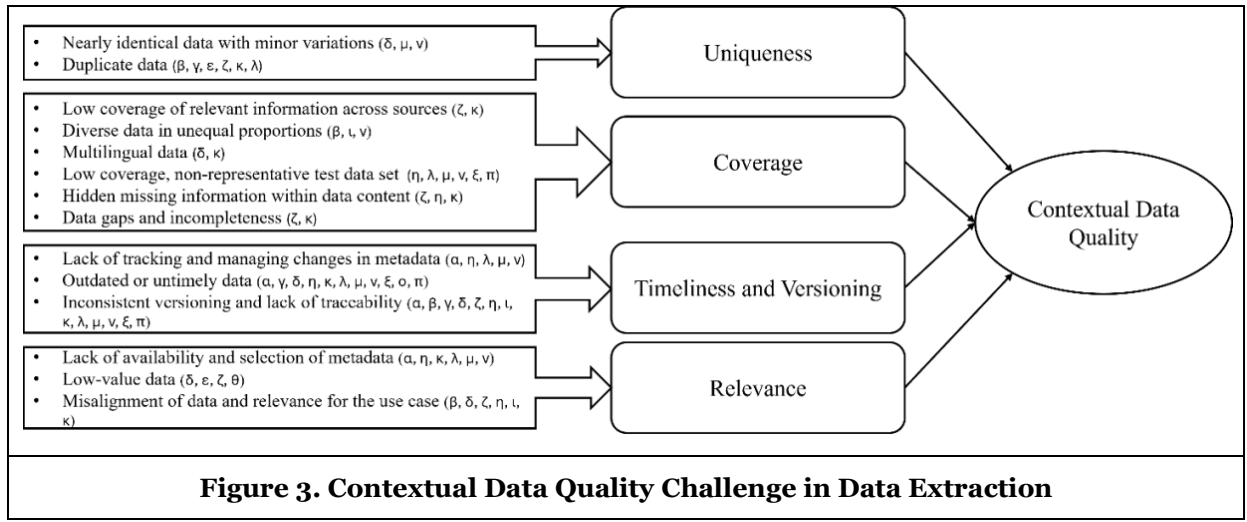

**Figure 3. Contextual Data Quality Challenge in Data Extraction**

**Uniqueness** refers to the distinctiveness of data and the absence of redundancies within the dataset. Nearly identical data with minor variations is especially challenging because it can introduce confusion and distort analytical outcomes according to ε. Duplicate adds to this complexity, leading to inefficiencies and potential misinterpretations in downstream processing. While minor variations can be valuable for tracking changes over time, such as exchange rates, exact duplications provide no additional information and should be eliminated. "Ideally you would have a good understanding of what information data covers but most time this is not the case, and you need a strategy to sort of disambiguate", κ mentioned.

**Coverage** concerns the extent to which data comprehensively represents the required information for a given use case. Low coverage across sources results in incomplete datasets. When data is diverse but unevenly distributed e.g., when certain categories or geographies are overrepresented, results can be distorted. Multilingual data adds another layer of complexity by challenging the consistency of interpretation. For instance, ν explains that diversification also matters during scale. "If we only work with German contracts, a German sample is sufficient. But to scale globally broader data is needed. Missing Middle East contracts mans that my conversation is incomplete".

**Timeliness and Versioning** relate to the timeliness and traceability of data over time. In real-time settings, it is critical to manage changes in metadata, maintain version control, and ensure data traceability. Outdated or untimely data can lead to obsolete conclusions, while inconsistent versioning complicates reproducibility. In this case it is especially important to define criteria for what determines a new version, β points out. Inconsistent versioning and missing traceability complicate the understanding of data evolution. These challenges arise particularly in dynamic and evolving analytical environments.

**Relevance** refers to the alignment of data with its intended purpose. Limited availability and selection of metadata can reduce the utility of extracted data. Low-value data that lacks contextual significance for the specific use case diminishes its applicability in practice. Furthermore, a misalignment between data content and the use case further diminishes the relevance of the extracted information. As λ states: "The data is relevant, if the user can relate it to the business problem".

## Representational Data Quality

Representational DQ pertains to the clarity and consistency of how data is represented. It focuses on whether data is formatted and structured in a way that supports its correct and effective interpretation and meaningful analysis. Two themes emerge from the analysis: Domain Knowledge and Representational Consistency (see Figure 4).

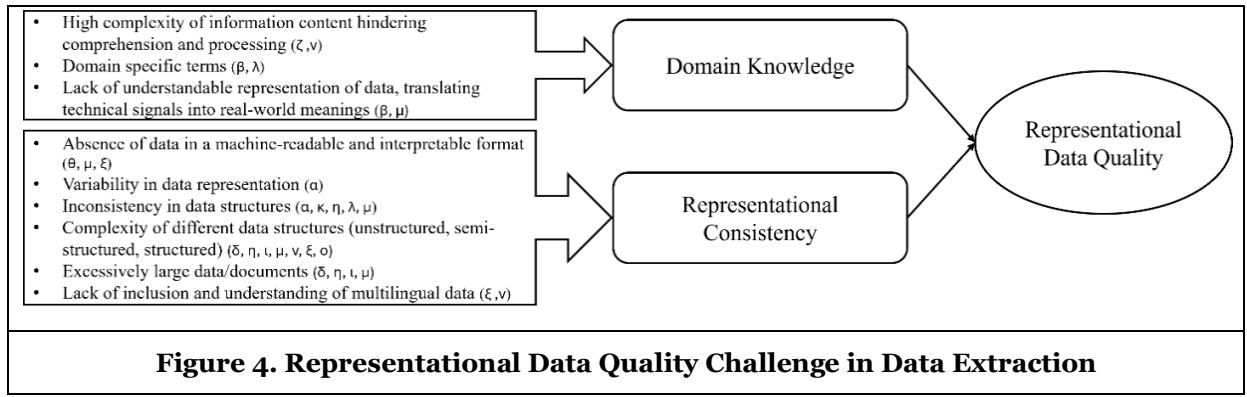

**Figure 4. Representational Data Quality Challenge in Data Extraction**

**Domain knowledge** refers to the understanding required to interpret complex and specialized information. When information is highly complex or filled with domain-specific terms, it can hinder comprehension, especially for non-expert users. Moreover, if the representation of data does not clearly translate technical representations into real-world meanings, interpretation becomes even more difficult. This reduces the clarity of extracted content and hinders effective decision-making. ι outlines his experience in which a message catalog was used to integrate domain knowledge: "The message catalog translates bus communications into real-world signals. For example, 0 means 'not charging', 1 means 'charging initiated', 2 means 'charging started'… The LLM must understand this domain knowledge."

**Representational consistency** describes the uniformity and clarity of data presentation. Data that is not in a machine-readable or standardized format complicates automated processing and integration. When representations vary across sources or structures are inconsistent, analytical coherence is undermined. The presence of multimodal data formats further increases this complexity. Excessively large data documents and a lack of inclusion and understanding of multilingual data add to these difficulties, possibly weakening the representational quality of the extracted data.

## Accessibility Data Quality

Accessibility DQ focuses on the ease with which data can be accessed, retrieved and integrated across systems. It ensures that data is readily available to authorized users when needed and can be processed effectively. The two themes manifesting in this category are Integration and Accessibility (see Figure 5).

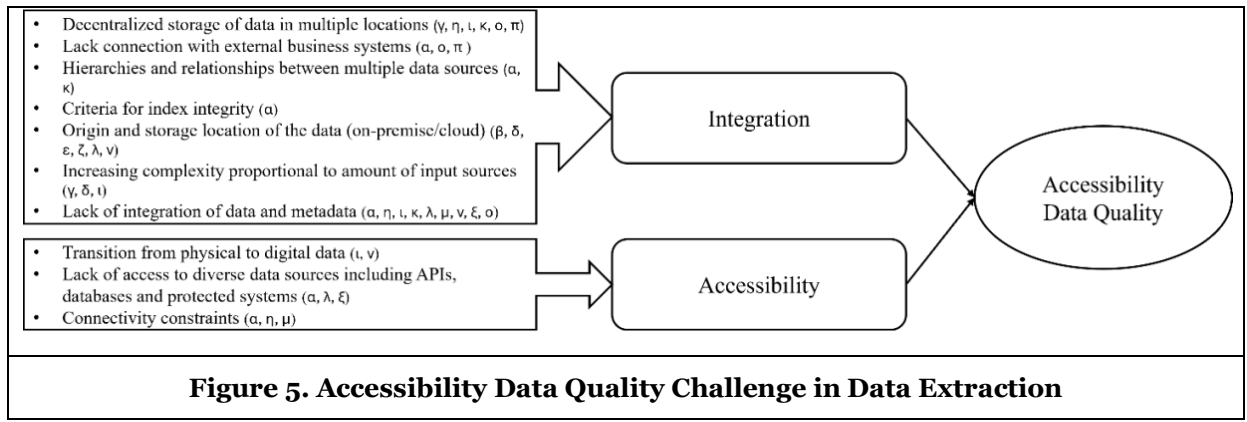

**Figure 5. Accessibility Data Quality Challenge in Data Extraction**

**Integration** refers to the ability to effectively combine, manage, and align data from multiple and often heterogeneous sources. When data is stored across multiple locations without centralized coordination, fragmentation of data increases. A lack of synchronization with external systems impedes the seamless flow of information across business functions. Hierarchies and complex relations between data sources demand robust strategies for integration. Moreover, ensuring the integrity of indexes and reconciling different storage infrastructures (e.g., on-premise vs. cloud) further complicates integration. As the number of input sources grows, the complexity of integrating them increases proportionally, particularly when metadata is not adequately merged.

**Accessibility** concerns the ease with which data can be accessed across various systems, from a technical perspective. The transition from physical to digital data can pose remarkable challenges in ensuring seamless access. Also, access can be constrained by infrastructural limitations, as µ explains: "Ensuring technical accessibility to diverse data sources is a key challenge. Integration often requires combining website data with SQL database results. Connectivity limitations and insufficient access rights often create barriers to efficient data extraction and processing, making accessibility a crucial prerequisite for any data-driven operation."

## Accountability Data Quality

Accountability DQ focuses on the assignment of responsibilities for specific data elements and the adherence to governance and compliance obligations. As shown in Figure 6, two themes are identified within this challenge: Ownership and Compliance.

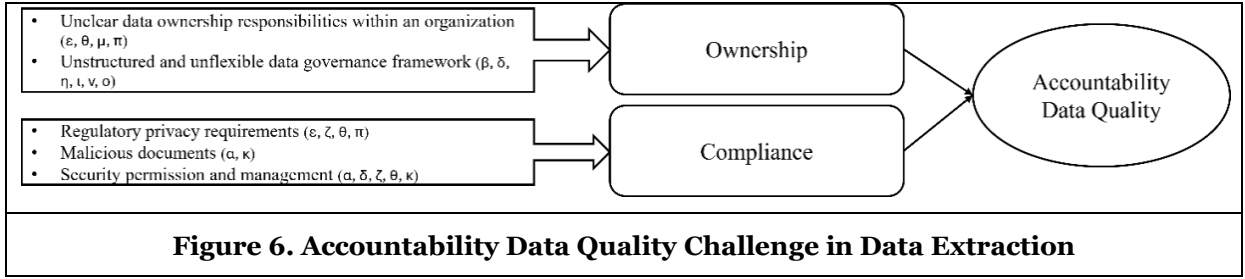

**Figure 6. Accountability Data Quality Challenge in Data Extraction**

**Ownership** refers to the clarity of responsibility for data within an organization. When responsibilities for DQ are not clearly assigned, uncertainty arises about who is accountable for maintaining and managing the data. This lack of clarity can result in gaps in oversight, delayed interventions, and reduced DQ. Moreover, an inflexible or poorly defined governance framework lacks the necessary adaptability to address evolving data demands.

**Compliance** concerns the organization's adherence to legal and regulatory-related requirements in data management. As regulatory standards tighten, organizations are required to implement robust measures for the secure handling of personal and sensitive data. Malicious documents, if not properly identified and managed, introduce significant cybersecurity threats. Furthermore, inadequate access control and poor management of security permissions can lead to unauthorized data exposure.

*Data Transformation*

The *Data Transformation* step processes the extracted data by segmenting it into manageable chunks, converting them into vector embeddings, and storing them in a vector database. This step emphasizes the semantic integration challenge of DQ, focusing on the coherent structuring of multimodal data, maintaining chunk integrity, preserving relationships and hierarchies, and ensuring the utility and contextual appropriateness of the transformed data.

## Semantic Integration

Semantic integration ensures that data from multiple sources is meaningfully combined, preserving its context and relationships to enable coherent understanding and analysis. Four second-order themes emerge from the analysis: Multimodality, Chunking, Relationships, and Information Utility (see Figure 7).

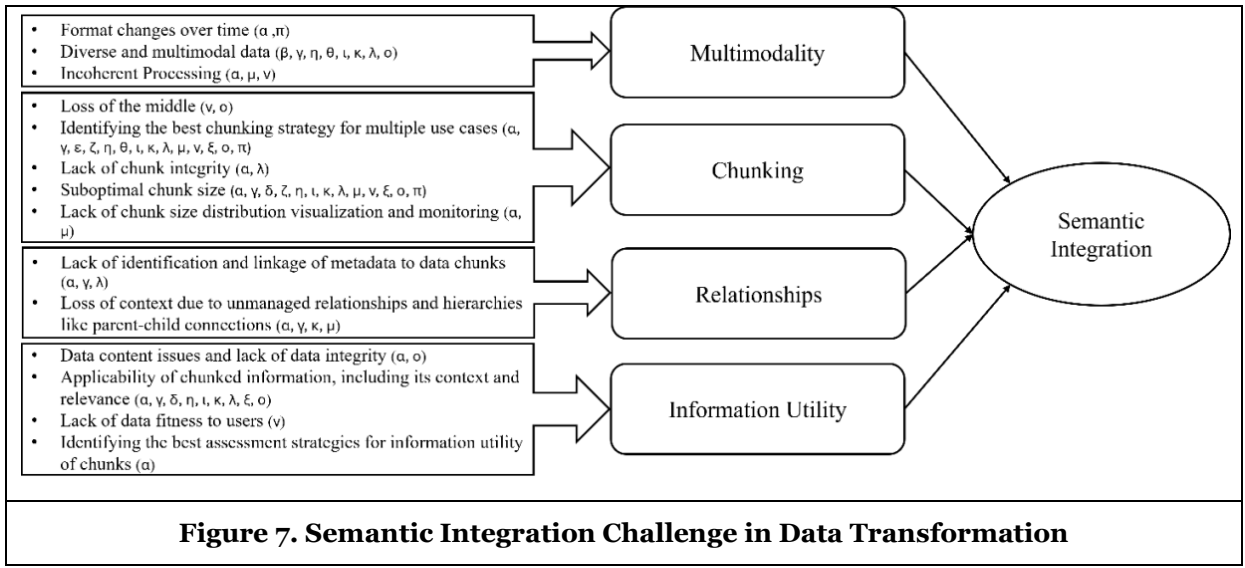

**Figure 7. Semantic Integration Challenge in Data Transformation**

**Multimodality** refers to the handling of diverse data formats and types during the transformation processing. Format changes, such as evolving file types or shifting structural standards, complicate the continuity of data handling and require frequent adaptations of processing pipelines. Multimodal sources such as the combination of text, tables, and images require tailored strategies that account for differing semantic representations to ensure coherent integration. Incoherent processing across formats disrupts the uniformity and consistency of the extracted information, which can lead to loss of meaning or analytical distortion in subsequent tasks.

**Chunking** involves slicing data into manageable units for processing and interpretation. A core challenge is preserving critical content within chunks, especially when information in the middle is overlooked. "When chunking is too large, such as feeding in a full page, models tend to focus on the beginning and end, often neglecting information in the middle. This 'lost in the middle' issue [...] can lead to missing the key answer, even if it's present in the chunk", ζ explains. The effectiveness of chunking also depends on choosing an appropriate strategy, i.e., where to chunk the available information. The selection of the chunking strategy must be context-sensitive, accounting for task-specific needs and data characteristics. Chunk integrity can be compromised when related information is split across different chunks, which can lead to semantic fragmentation.

**Relationships** describe the contextual connections between data chunks and the preservation of their interdependencies. When metadata is not properly linked to corresponding chunks, the information becomes harder to interpret and to verify the transparency of the data. Additionally, the absence of managed structural relationships, such as parent-child connections in hierarchical data, can lead to a breakdown of logical flow and interdependency, limiting the semantic completeness of the dataset.

**Information Utility** concerns the relevance, applicability, and fitness of chunked data for its intended purpose. Data content issues such as missing or inconsistent information reduce overall data integrity while the applicability of chunked information depends on whether the content retains its original context and relevance for the task at hand. Poorly structured chunks may contain low-value or irrelevant content, diminishing their usefulness. The data must also be evaluated in terms of its fitness for the user, meaning it should align with user needs, expectations, and domain requirements.

### *Prompt & Search*

During *Prompt & Search*, user input is transformed into an embedding and matched against the company specific information to retrieve the most relevant information. This phase engages with interpretability and provenance DQ challenges.

## Interpretability

Interpretability refers to the degree to which data and outputs can be easily understood, logically traced and meaningfully applied when searching for the relevant information. This ensures that the extracted information aligns with the user's intent and supports reliable downstream use. As shown in Figure 8, we identify wo themes: Intent Recognition and Context Selection.

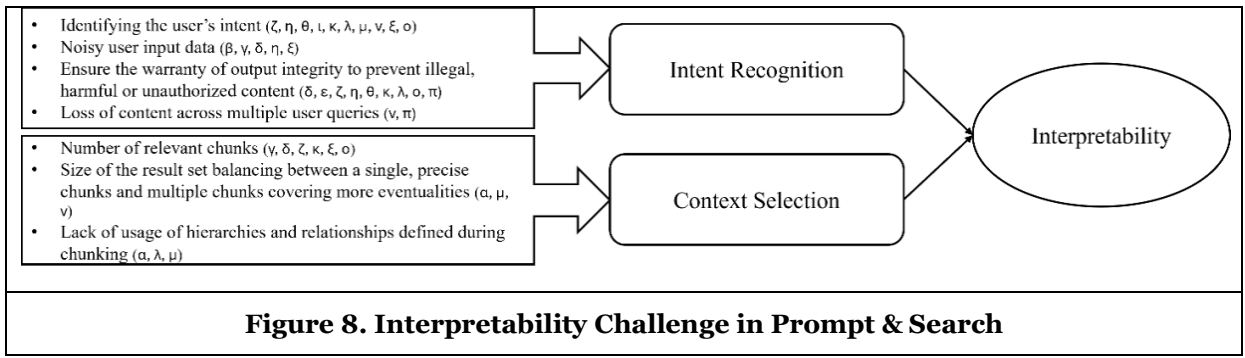

**Figure 8. Interpretability Challenge in Prompt & Search**

**Intent Recognition** refers to the ability to correctly understand the user's real intent. Understanding the intent becomes challenging when user inputs are vague, fragmented, or inconsistent, often requiring systems to infer meaning from limited signals. Noisy input, such as typographical errors or irrelevant phrasing, further reduces the clarity of interpretation and increases the risk of misalignment. Safeguarding output integrity is also crucial to prevent the generation of responses that may be illegal, harmful, or unauthorized.

**Context Selection** describes retrieving the most relevant information chunks in response to a user query. Selecting too few chunks may lead to incomplete results, while selecting too many risks overwhelming the model with irrelevant data. The balance between precision and coverage is a recurring trade-off. While smaller chunks may isolate key answers, larger chunks preserve context but risk burying important details. α explains, "Ideally, you would return just one perfect chunk with the exact answer. But in practice, you often need multiple chunks to cover enough context." Additionally, failing to utilize previously defined chunk hierarchies and semantic relationships can result in incoherent or disjointed information affecting the interpretability of the results, as γ illustrates.

## Provenance

Provenance focuses on maintaining the integrity and traceability of data throughout its lifecycle. It ensures transparency regarding the origins of data, how it has been processed, and how it is accessed and used. Figure 9 depicts the four themes we identified: Integration, Precision, Traceability, and Security.

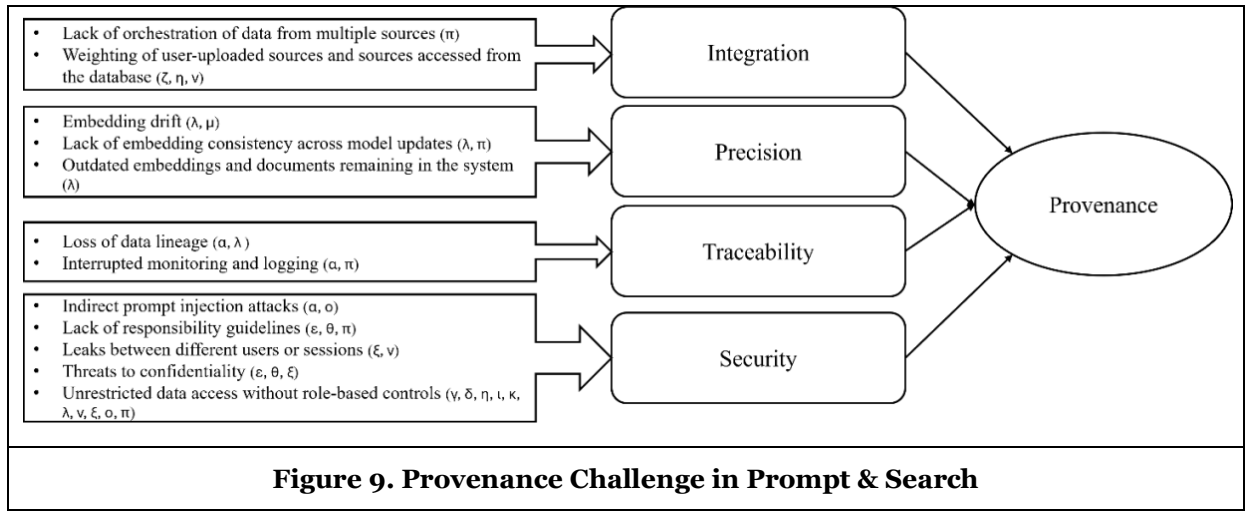

**Figure 9. Provenance Challenge in Prompt & Search**

**Integration** refers to the orchestration of data from multiple, heterogeneous sources. When inputs from different origins, e.g. user-uploaded sources and data accessed from databases, are not properly coordinated, the resulting data can be fragmented or redundant. Moreover, imbalances in the implicit weighting of certain sources can introduce unintended biases, ultimately leading to distorted or incomplete results.

**Precision** addresses the relevance and temporal currency of data. Embedding drift occurs when the vector embeddings used for retrieval change over time or between model versions, causing queries and documents to no longer align semantically. This misalignment degrades retrieval quality by making vector representations incompatible or less reliable and can lead to irrelevant or hallucinatory outputs in a RAG system. A lack of consistency across model updates further complicates retrieval, as embedding logic may differ across versions. The presence of outdated documents or embeddings in the system can also interfere with relevance, especially if these are not flagged or internally linked.

**Traceability** describes the ability to reconstruct the origin, transformation, and use of data within the system. The loss of data lineage can impede understanding of where data originated and how it has been processed. Inconsistent or missing monitoring data impedes the tracing of changes and decisions, especially in complex multi-step extraction processes. π describes traceability as a cascading challenge throughout the system that influences downstream processes, especially when DQ must be verified retrospectively.

**Security** involves safeguarding data from unauthorized access, manipulation and leakage as well as ensuring its integrity. Indirect prompt injection attacks can exploit model behavior to expose sensitive content. Session-based data leaks where information unintentionally flows between users pose a significant risk in shared environments. Threats to confidentiality are amplified when sensitive data is used in retrieval without constraints. The absence of role-based access control further heightens these risks by failing to restrict data visibility based on user roles. θ depicts access control as the trickiest part of the system because "…the main risk lies in the model's vulnerability to leaking sensitive information during generation if retrieval is not properly scoped."

### *Generation*

During *Generation*, retrieved chunks are combined with the predefined system prompt and passed to the LLM to generate an answer to the user's prompt. This step creates Conversational Style and Credibility challenges of DQ.

## Conversational Style

Conversational style characterizes the way information is presented in interactions. This dimension ensures that answers are not only accurate in content but also accessible and aligned with human communication expectations. Two themes are central to this: Format Adherence and Linguistic Coherence (see Figure 10).

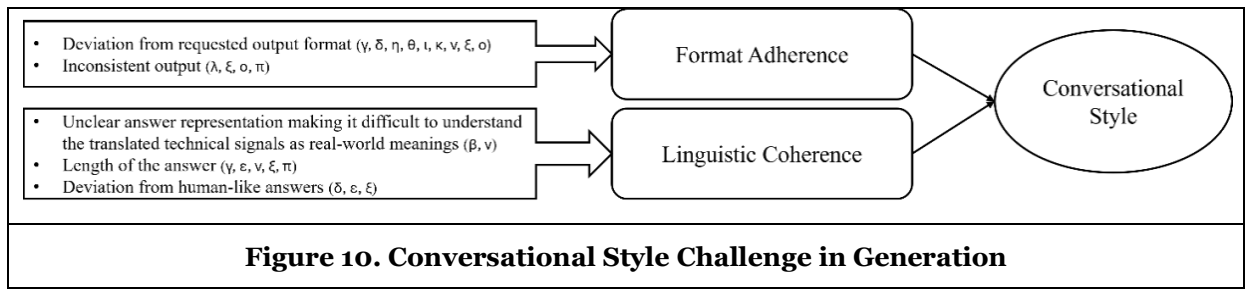

**Figure 10. Conversational Style Challenge in Generation**

**Format Adherence** relates to the extent to which outputs consistently follow the requested format in the response. When language models fail to follow structured formatting instructions, the outcome becomes difficult to parse or process automatically. Deviations from the requested output format can lead to misunderstandings and misinterpretations. Inconsistent formatting not only disrupts automated pipelines but also reduces trust in the model's reliability. This requires additional validation and correction steps to align the results with intended use cases. When the structure or semantics of the answer are ambiguous, users struggle to translate technical or encoded outputs into real-world meanings.

**Linguistic Coherence** concerns the clarity, readability, and naturalness of responses. Responses that are too short may neglect essential details, while overly long one's risk burying key information or becoming difficult to follow. Additionally, outputs that deviate too far from natural human expression can appear robotic or artificial. This undermines user confidence and reduces the conversational quality of the interaction.

## Credibility

Credibility pertains to the trustworthiness and reliability of data, ensuring that outputs are accurate, valid, and suitable for their intended purpose. It is central to user acceptance and decision-making, as users must be able to rely on information produced. This challenge becomes particularly relevant in high-stakes contexts, where DQ has direct implications on decision making. Two key themes define this dimension as shown in Figure 11: Correctness and Completeness.

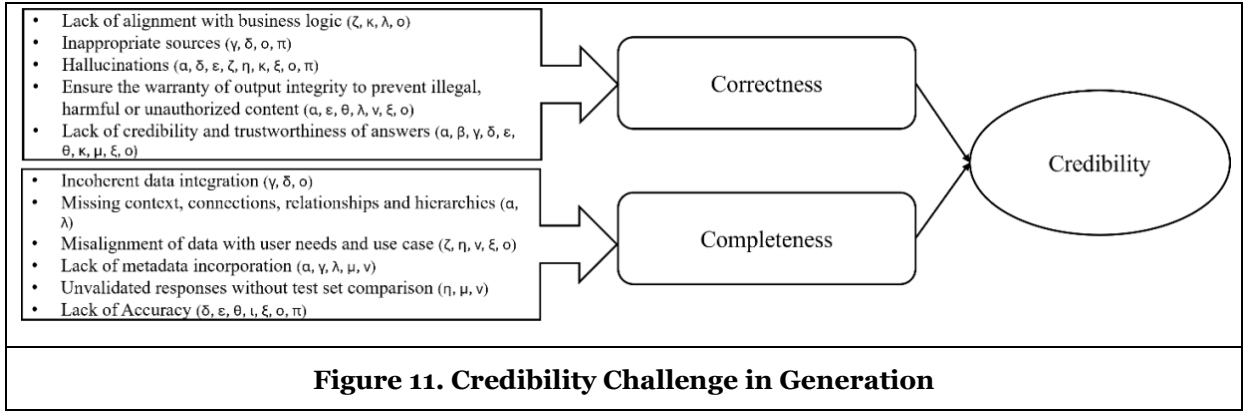

**Figure 11. Credibility Challenge in Generation**

**Correctness** explains the alignment of extracted data with factual and contextual accuracy. When outputs fail to reflect business logic, such as producing technically correct but contextually irrelevant answers, the value diminishes. The use of inappropriate or low-quality sources can introduce bias or misinformation. Hallucinations, where the model fabricates content not grounded in the source material, reduce trust in the LLM. Safeguarding integrity by avoiding displaying undesirable, generated content is essential to prevent harmful, or illegal outputs. A lack of transparency in how answers are generated leads to reduced credibility. As η observes providing source references is essential for credibility and user acceptance.

**Completeness** concerns the extent to which data captures all information. When data is poorly integrated, information may appear fragmented or disjointed, reducing semantic coverage. Missing contextual links, such as connections, relationships, or hierarchies, further weaken the completeness of the output. From a user perspective, the data is often perceived as insufficient if it does not meet user's expectations or aligns with the intended use case.

To summarize, Figure 12 illustrates how the challenges identified in this study are aligned with the established steps of RAG. Each dimension is mapped to a distinct phase based on where its influence and challenges occur. This mapping highlights that DQ in RAG cannot be addressed by a static but dynamic, step-aware approach. It requires tailored attention at each step of RAG to ensure end-to-end coherence.

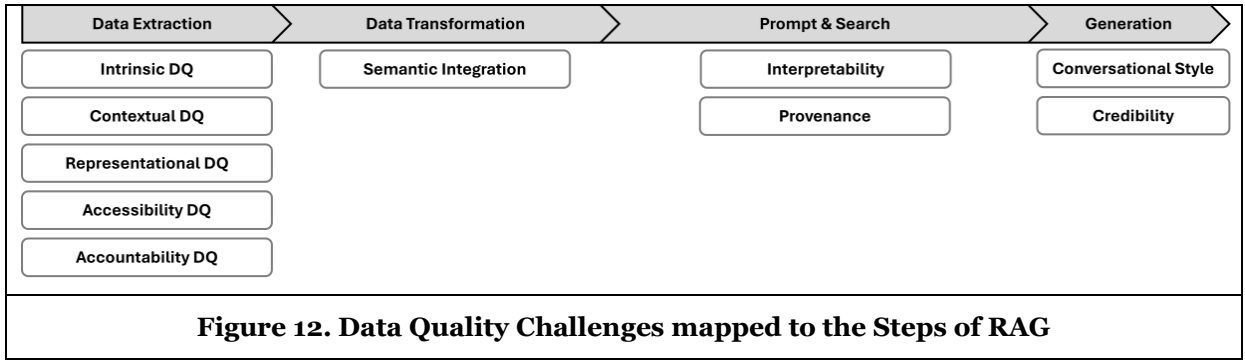

**Figure 12. Data Quality Challenges mapped to the Steps of RAG**

# Discussion

This study systematically examines DQ challenges across different steps of RAG through exploratory expert interviews. Thereby, we identify 26 DQ challenges distributed over the different processing steps of RAG. Our findings reveal three key insights about DQ in RAG systems: the emergence of the intersection between DQ and data governance with accountability as an additional challenge going beyond established DQ framework, the focus of challenges in the early processing steps of RAG systems, and the complex temporal dependencies that exist across the entire RAG system requiring step-specific assessment.

## Implications for Theory

We contribute to the DQ literature by introducing *Accountability* as an additional dimension for RAG systems, thereby extending Wang and Strong's (1996) foundational DQ framework. While traditional DQ dimensions remain relevant, particularly at the *Data Extraction* step, RAG systems require explicit attention to ownership and compliance challenges that emerge from integrating continuously evolving organizational knowledge. In contrast to static database environments where existing DQ constructs were developed (Chiang & Miller, 2008; Yeh & Puri, 2010), RAG systems create traceability challenges that expose critical gaps in established governance frameworks. Traditional stewardship models define clear role boundaries between data owners, data stewards, and data custodians (Khatri & Brown, 2010), but these roles assume direct oversight of immediate use of data assets. However, once documents are transformed into vector embeddings and stored in databases, traditional stewardship mechanisms lose visibility into which specific sources influence RAG outputs. Our participants revealed that outdated data may persist in vector databases even after source documents are updated, yet governance frameworks lack mechanisms for tracing which embedded representations are actively used during generation. When data stewards cannot identify which outdated information influences business decisions through persistent vectorstores, traditional accountability mechanisms fail. This reveals fundamental limitations in existing stewardship models that were designed for direct data oversight rather than the indirect, transformed data use characteristic of RAG systems. While this *Accountability* dimension aligns with recent IS research emphasizing governance in complex data ecosystems (Al Wahsi et al. (2022) and Karkošková (2023)), our findings extend these frameworks by demonstrating how RAG systems require governance models capable of maintaining data lineage across vectorization processes. We thus argue that Accountability represents an emergent DQ requirement necessitating new stewardship approaches for enterprise AI systems.

Next, our findings point to a strong "front-loading" of DQ challenges: the Data Extraction step alone accounts for 73 coded challenges, substantially more than the 15–24 coded challenges identified in downstream steps This builds upon existing IS theory emphasizing the importance of data provenance and data valuation to only select relevant data for the training of effective AI systems (Schneider et al., 2023), but adds a process-aware lens: initial quality challenges are amplified through RAG's sequential pipeline. Consequently, proactive quality assurance is needed to ensure sufficient performance in downstream steps, echoing discussions around data-centric *AI* (Jakubik et al., 2024). Finally, the complex temporal dependencies we identify challenge dominant assumptions in IS literature that treat DQ as a multi-dimensional but static construct (e.g., (Yu et al., 2024)). In contrast, RAG systems exhibit dynamic propagation patterns in which the quality of outputs at each step depends on cumulative upstream conditions. Our results suggest that enterprises must abandon monolithic DQ assessments in favor of step-specific interventions that account for how quality issues unfold and interact over time within RAG systems.

## Implications for Practice

For practitioners, this work identifies and synthesizes DQ challenges across the RAG pipeline and provides guidance for addressing them. The pronounced front-loading of challenges highlights the critical importance of early-stage interventions. Implementing rigorous initial data extraction and validation mechanisms can significantly reduce downstream quality issues and associated corrective costs, aligning with Redman's (2017) observations on proactive quality management. While full-step monitoring offers high reliability, it may be costly. Resource-constrained organizations can prioritize checkpoints at critical stages or use lightweight quality controls like sampling or anomaly detection. Future work should explore scalable and cost-effective monitoring strategies tailored to different organizational capacities.

## Limitations and Future Work

Our research comes with limitations that offer areas for further research: First, while our qualitative expert interviews provide rich insights, this approach lacks quantitative validation of the relative impact and frequency of different DQ challenges. Future research should develop quantitative methods and metrics to systematically measure DQ. RAGAs (Es et al., 2024) presents a promising starting point by offering evaluation of RAG outputs along challenges such as context relevance and answer faithfulness. However, recent research demonstrates mixed results with this framework, where LLM-based evaluation often yielded different results compared to expert assessment (Holstein et al., 2025). With the further development of frameworks like RAGAs, researchers could quantitatively link DQ challenges to measurable declines in RAG effectiveness. Second, our findings are based on insights from practitioners at only two large IT service companies, potentially limiting generalizability. Future research should confirm whether these DQ challenges are consistent across a broader range of organizations and industries, or if additional challenges emerge in different contexts. Third, this study examined DQ challenges within individual RAG processing steps. However, this approach does not capture the potential relationships and interactions between challenges across different steps, limiting our understanding of how DQ challenges are propagated through RAG systems. Yet, understanding how the identified DQ challenges in one step influence challenges in the following steps could provide valuable insights. Future work should, therefore, investigate these cross-dimensional relationships to identify dependencies where DQ issues are created and then propagate through multiple processing stages throughout the RAG system. Lastly, while our research comprehensively identifies challenges, it does not systematically evaluate solutions or mitigation strategies for the identified DQ issues. Future work should transition from identifying problems to implementing and assessing solutions for DQ management in RAG. Through addressing our limitations, future research can build upon our framework to develop more comprehensive, validated, and actionable approaches to managing DQ in enterprise RAG implementations.

## Conclusion

This study systematically examines how DQ challenges emerge across RAG through 16 expert interviews with professionals from leading multinational IT service firms. We map 26 DQ challenges to the four sequential RAG steps, *Data Extraction*, *Data Transformation*, *Prompt & Search*, *Generation*, providing empirical evidence of step-specific DQ dynamics in enterprise RAG systems. Our findings introduce a step-aware perspective on DQ, demonstrating that quality issues do not emerge uniformly but propagate and transform along the distinct RAG steps. Furthermore, we extend existing DQ frameworks by revealing that while established dimensions such as intrinsic DQ dominate in early steps, new challenges such as semantic integration dynamically arise during downstream processing. Together, these insights highlight that established holistic approaches to DQ are insufficient for RAG systems. Managing DQ for these novel systems requires proactive, step-specific interventions across the full pipeline. As enterprises increasingly embed LLMs into business-critical workflows, ensuring DQ at every step of RAG becomes a strategic imperative for building reliable, trustworthy, and competitive AI systems.